%% file: main.tex
\documentclass{article}

  \usepackage[preprint, nonatbib]{neurips_2026}

  \usepackage[utf8]{inputenc} 
  \usepackage[T1]{fontenc}    
  \usepackage{hyperref}       
  \usepackage{url}            
  \usepackage{booktabs}       
  \usepackage{amsmath}        
  \usepackage{amsfonts}       
  \usepackage{nicefrac}       
  \usepackage{microtype}      
  \usepackage{xcolor}         
  \usepackage{graphicx}
  \usepackage{float}
  \usepackage{booktabs}
\usepackage{subcaption}

\newcommand{\myparagraph}[1]{\smallskip\noindent\textbf{#1}}

  \title{PathReportEval: A Systematic Benchmark for Pathology Report Generation}

  \author{%
  Suryakant Singh \\
  Department of Biomedical Informatics\\
  Stony Brook University\\
  Stony Brook, NY, USA \\
  \texttt{suryakant.singh@stonybrook.edu} \\
  \And
  Sejuti Majumder \\
  Department of Biomedical Informatics\\
  Stony Brook University\\
  Stony Brook, NY, USA \\
  \texttt{sejuti.majumder@stonybrook.edu} \\
  \And
  Beatrice Knudsen \\
  Department of Pathology\\
  University of Utah\\
  Salt Lake City, UT, USA \\
  \texttt{beatrice.knudsen@path.utah.edu} \\
  \And
  Joel Saltz \\
  Department of Biomedical Informatics\\
  Stony Brook University\\
  Stony Brook, NY, USA \\
  \texttt{joel.saltz@stonybrook.edu} \\
  \And
  Prateek Prasanna \\
  Department of Biomedical Informatics\\
  Stony Brook University\\
  Stony Brook, NY, USA \\
  \texttt{prateek.prasanna@stonybrook.edu} \\
  }

\begin{document}

  \maketitle

\begin{abstract}
Pathology report generation from whole-slide images (WSIs) is a rapidly growing multimodal learning problem, yet progress is difficult to measure because existing studies use heterogeneous datasets, model settings, visual encoders, and evaluation protocols. Moreover, commonly used natural language generation metrics, including BLEU, ROUGE, and METEOR, primarily reward lexical similarity and can fail to detect clinically consequential errors such as omitted diagnoses, hallucinated findings, or discordant tumor attributes.

We present a standardized benchmark and evaluation framework for pathology report generation. The benchmark evaluates four representative generation methods across three datasets, TCGA, HistAI, and REG 2025, using three pathology foundation encoders, CONCHv1.5, UNI2-h, and H-Optimus-1. Our framework standardizes the full experimental pipeline, including preprocessing, tokenization, feature extraction, training, decoding, and evaluation, enabling fair comparison across model architectures and visual representations. The framework is modular and extensible, allowing new methods, datasets, and encoders to be incorporated under a shared protocol.

A central contribution of this work is the Clinical Report Quality Score (CRQS), a clinically grounded metric for evaluating factual correctness in generated pathology reports. CRQS maps reference and generated reports into structured clinical attributes and measures four complementary dimensions of report quality: clinical fact coverage, key information recall, hallucination rate, and clinical discordance. This decomposition provides both an aggregate score and interpretable sub-scores that expose clinically meaningful failure modes.

Our experiments show that conventional language-generation metrics are weakly aligned with clinical correctness and can overestimate report quality. CRQS, in contrast, identifies substantial differences between models and encoders that are not captured by lexical metrics. By providing a standardized benchmark, a public plug-and-play framework, and a structured clinical evaluation metric, this work establishes a reproducible foundation for measuring progress in pathology report generation. Evaluation framework: \hbox{https://github.com/surykntsingh/PathBench}
\end{abstract}
  \input{sections/introduction}

  \input{sections/related_works}
  \input{sections/benchmark_and_evaluation_framework}
  \input{sections/experiments}
\input{sections/analysis}
  \input{sections/discussions_and_conclusion}
\begin{ack}
This research was partially supported
by NIH grants 1R01CA297843-01, 1R03DE033489-01A1,
and NSF grant 2442053. The content is solely the responsibility of the authors and does not necessarily represent the
official views of the NIH.
\end{ack}

\clearpage
    \bibliographystyle{IEEEtran}
\bibliography{refs}

  \clearpage
  \input{sections/appendix}
  

  \end{document}

%% file: sections/introduction.tex
\section{Introduction}

Pathology report generation from whole-slide images (WSIs) is an emerging multimodal learning task that requires models to convert gigapixel histopathology images into clinically meaningful textual reports ~\cite{lucassen2025pathology, hu2025pathology,jin2026grounded,gao2025s2d}. Recent advances in vision--language modeling and pathology foundation models have made this task increasingly feasible\cite{conch,uni,hoptimus1}, enabling systems that can integrate visual evidence across large tissue regions and generate diagnostic descriptions. However, progress in this area remains difficult to measure. Existing studies ~\cite{histgen, wsi_caption,bigen, scout} often use different datasets, preprocessing pipelines, visual encoders, model implementations, decoding strategies, and evaluation protocols, making direct comparison across methods unreliable. As a result, it is often unclear whether reported performance gains reflect improvements in model design or differences and biases in experimental setup.

A second challenge is evaluation. Most pathology report generation studies rely on standard natural language generation (NLG) metrics, such as BLEU~\cite{bleu}, ROUGE~\cite{rouge}, and METEOR~\cite{meteor}. These metrics measure lexical overlap between generated and reference reports, but they do not directly assess whether clinically important information is preserved~\cite{yu2023radcliq,baharoon2026crimson}. In pathology, this distinction is critical: a generated report may be lexically similar to a reference report while omitting a key diagnosis, misstating tumor grade or stage, hallucinating lymph-node involvement, or contradicting margin status. Such errors are not merely linguistic failures; they reflect failures of clinical fidelity. Thus, evaluating pathology report generation requires metrics that move beyond surface-form similarity and explicitly measure the correctness of clinically meaningful content.

\begin{figure*}[t]
\centering
\includegraphics[width=\textwidth]{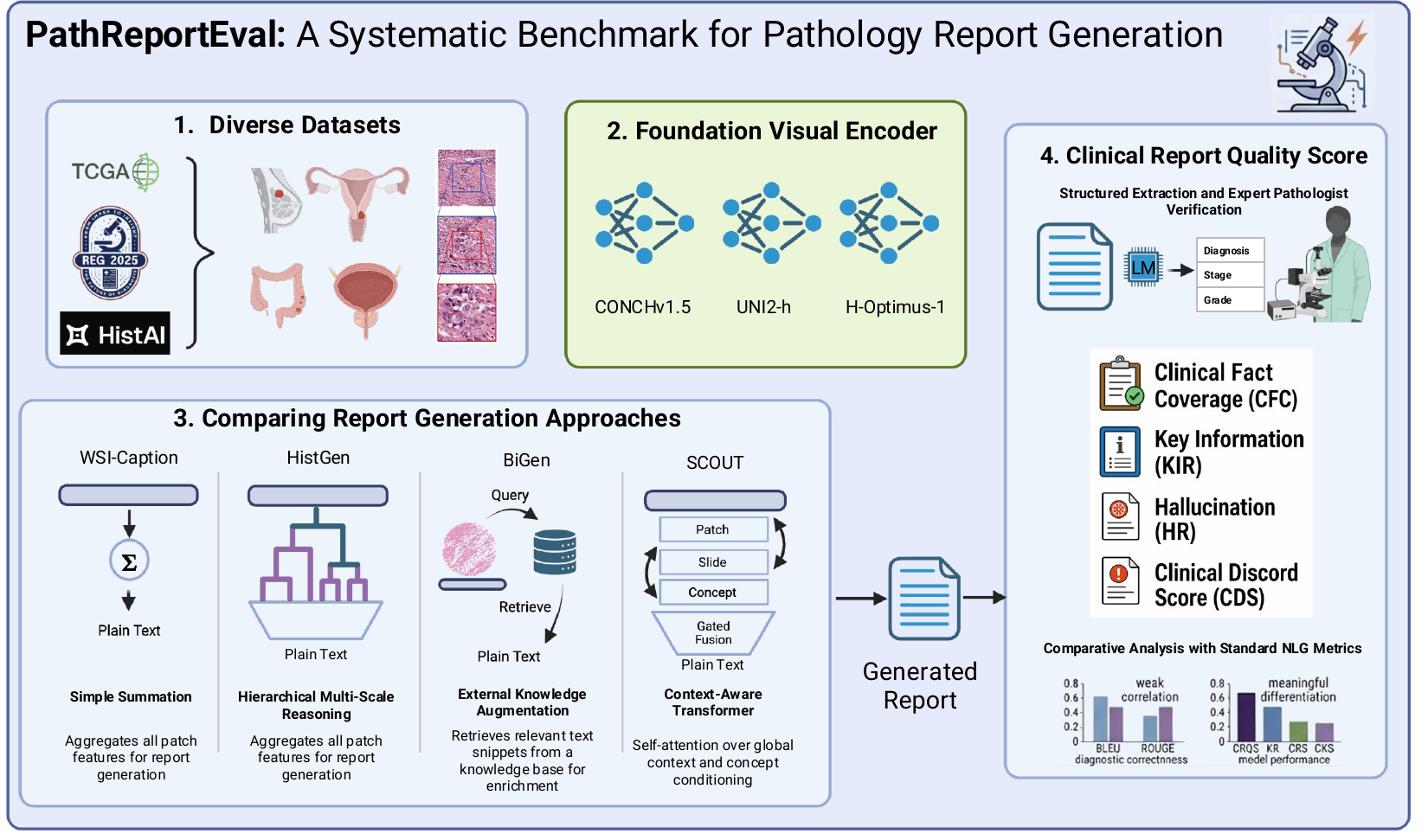}
\caption{Overview of the proposed pathology report generation benchmark and evaluation framework. The framework standardizes data splits, feature extraction, model training, and evaluation across datasets, encoders, and report generation methods, while also incorporating the clinically grounded CRQS pipeline for structured assessment of report quality.}
\label{fig:benchmark_overview}
\end{figure*}

In this work, we introduce a standardized benchmark and evaluation framework for pathology report generation from WSIs. We evaluate four state-of-the-art report generation methods across three datasets with distinct reporting characteristics and three pathology visual encoders representing different foundation-model families. By evaluating all methods under a shared experimental protocol, our benchmark enables systematic comparison across model architectures, datasets, and visual representations. To support reproducibility and future extension, we further provide a modular plug-and-play framework that standardizes preprocessing, tokenization, feature extraction, training, decoding, and evaluation. The framework is designed to allow new report generation models, datasets, and visual encoders to be incorporated with minimal changes while preserving a common evaluation pipeline.

To address the limitations of lexical evaluation, we propose the \emph{Clinical Report Quality Score} (CRQS), a pathologist-verified structured metric for assessing clinical fidelity in generated pathology reports. CRQS is motivated by synoptic reporting practices, such as those recommended by the College of American Pathologists~\cite{capCancerProtocols,CAP_Breast_2023}, in which key diagnostic and prognostic attributes are documented using standardized fields. Rather than comparing reports only as free text, CRQS maps both generated and reference reports into structured clinical attributes, such as diagnosis, histologic type, tumor grade, stage, margin status, lymph-node involvement, biomarker status, and other dataset-specific pathological findings. It then evaluates report quality using four interpretable components: clinical fact coverage, key information recall, hallucination rate, and clinical discordance. This decomposition allows CRQS to capture distinct failure modes, including omitted findings, missing high-impact diagnostic fields, unsupported generated facts, and direct contradictions with the reference report.

Across datasets, methods, and encoders, our analysis shows that conventional NLG metrics can obscure clinically meaningful differences between systems. Models with similar lexical scores may differ substantially in their ability to recover key diagnostic information, avoid hallucinations, and maintain consistency with reference clinical attributes. CRQS provides a complementary evaluation signal that is more directly aligned with clinical content and supports fine-grained error analysis. Together, the benchmark, public framework, and CRQS provide a reproducible foundation for measuring progress in pathology report generation. Our contributions are threefold:
\begin{enumerate}
    \item We introduce a \textbf{standardized benchmark for pathology report generation} from WSIs, evaluating representative methods across multiple datasets and pathology visual encoders under a shared experimental protocol.
    \item We provide a \textbf{modular plug-and-play benchmarking framework} that standardizes preprocessing, feature extraction, training, decoding, and evaluation, enabling fair comparison and rapid extension to new models, datasets, and encoders.
    \item We \textbf{propose CRQS, a clinically grounded evaluation metric} that assesses generated reports using structured clinical attributes and decomposes report quality into fact coverage, key information recall, hallucination, and clinical discordance.
\end{enumerate}

Figure~\ref{fig:benchmark_overview} presents an overview of this benchmark and evaluation pipeline, highlighting how datasets, visual encoders, report generation models, and clinical evaluation components are organized within a single standardized framework. All code related to this evaluation framework is available at \hbox{https://github.com/surykntsingh/PathBench}

%% file: sections/related_works.tex
\section{Related Work}

\paragraph{Pathology report generation from WSIs.}
Automated pathology report generation has recently emerged as a challenging multimodal learning
problem, requiring models to translate gigapixel whole-slide images (WSIs) into clinically meaningful 
text~\cite{gamper2021multiple,huang2023visual,lu2023visual}. Early approaches extended image captioning frameworks to WSIs using multiple instance learning
(MIL), where patch-level features are aggregated to produce slide-level reports. WSI-Caption
(MI-Gen)~\cite{wsi_caption} demonstrated the feasibility of transformer-based report generation at gigapixel scale.
Subsequent methods have introduced more structured modeling strategies: HistGen~\cite{histgen} uses local--global
hierarchical encoding and cross-modal context interaction to better align multi-scale visual evidence
with textual descriptions, while BiGen~\cite{bigen} incorporates retrieval-based knowledge to enrich visual
representations with semantically relevant information. These works represent important progress, but
they are typically evaluated under heterogeneous settings, including different datasets, preprocessing
pipelines, visual encoders, decoding strategies, and evaluation protocols. As a result, it remains
difficult to determine whether observed performance differences arise from model design or from
inconsistent experimental conditions.

\paragraph{Vision--language models and pathology foundation encoders.}
Recent advances in vision--language learning have produced powerful pretrained encoders for
histopathology, including CONCH~\cite{conch}, UNI~\cite{uni}, and H-Optimus~\cite{hoptimus1}. These models provide strong visual
representations and have improved performance across a range of pathology tasks. However, most
work using such encoders focuses on representation learning or downstream discriminative tasks,
rather than systematic evaluation of generative pathology reporting. Moreover, report generation
requires not only visual recognition but also clinically faithful selection, organization, and expression
of diagnostic findings. Thus, there is a need to evaluate how different visual encoders influence
clinical correctness in generated reports under a common benchmark.

\paragraph{Evaluation of medical report generation.}
Evaluation remains a major bottleneck for medical report generation. Most existing studies rely on
standard natural language generation metrics such as BLEU~\cite{bleu}, ROUGE~\cite{rouge}, and METEOR~\cite{meteor}, which measure
lexical overlap between generated and reference reports. While these metrics are easy to compute,
they do not directly assess factual correctness or clinical relevance. A generated report can achieve
high lexical similarity while omitting key diagnostic findings, introducing unsupported clinical
attributes, or contradicting the reference diagnosis. Prior work in radiology has explored structured
entity-based or graph-based evaluation~\cite{jain2021radgraph,yu2023radcliq,baharoon2026crimson}, and recent pathology report generation work has introduced
fact-based metrics such as FACTent~\cite{miura-2021-improving, wsi_caption}. These efforts move beyond surface-form similarity, but they
typically treat facts as unstructured entities and do not explicitly model the structured nature of
pathology reporting.

\paragraph{Need for standardized and clinically grounded benchmarking.}
Benchmarking has been central to progress in computational pathology, particularly for classification,
segmentation, retrieval, and survival prediction. However, existing pathology benchmarks largely
focus on visual prediction tasks and do not address multimodal report generation. Current report
generation methods are therefore evaluated in isolation, limiting reproducibility and fair comparison.
In this work, we address this gap by introducing a standardized benchmark across models, datasets,
and visual encoders, together with a modular plug-and-play framework for reproducible evaluation.
We further propose the Clinical Report Quality Score (CRQS), a structured metric that evaluates
generated reports using clinically meaningful fields aligned with synoptic reporting practice. Unlike
lexical metrics or unstructured fact-overlap scores, CRQS decomposes report quality into clinical fact
coverage, key information recall, hallucination rate, and clinical discordance, enabling both aggregate
comparison and interpretable error analysis.

%% file: sections/benchmark_and_evaluation_framework.tex
\section{Benchmark and Evaluation Framework}
\label{sec:benchmark}
\begin{table*}[t]
\centering
\caption{\textbf{Characteristics of datasets used for pathology report generation.}
The datasets differ substantially in size, institutional diversity, annotation noise, and report structure, enabling a comprehensive evaluation of model robustness and generalization.}
\label{tab:dataset_comparison}
\begin{tabular}{l|c|c|c|c}
\toprule
\textbf{Dataset} & \textbf{Train / Val / Test} & \textbf{Diversity} & \textbf{Report Homogeneity} & \textbf{Annotation Noise} \\
\midrule
TCGA  & 7074 / 884 / 885  & Low & Medium & High \\
HistAI & 9980 / 1248 / 1248  & High & Low & Medium  \\
REG 2025 & 5924 / 740 / 741  & High & High & Low  \\
\bottomrule
\end{tabular}
\end{table*}

\subsection{Task Definition}

We study pathology report generation from whole-slide images (WSIs). Given a WSI $X$, a model
generates a report $\hat{R} = (r_1, r_2, \ldots, r_T)$, where each token $r_t$ is drawn from a fixed
vocabulary. The corresponding reference report is denoted by $R$. Unlike generic image captioning,
this task requires models to recover diagnostically relevant visual evidence, express it in clinically
appropriate language, and preserve structured attributes such as diagnosis, grade, stage, margin status,
and other pathology-specific findings.

Our goal is not to introduce a new report generation architecture, but to evaluate existing systems
under a controlled and reproducible benchmark. We therefore define a standardized protocol for
comparing models across datasets and visual encoders, using both conventional language-generation
metrics and the proposed clinically grounded CRQS metric.

\begin{figure}[t]
    \centering
    \begin{minipage}[t]{0.49\linewidth}
        \centering
        \includegraphics[width=\linewidth]{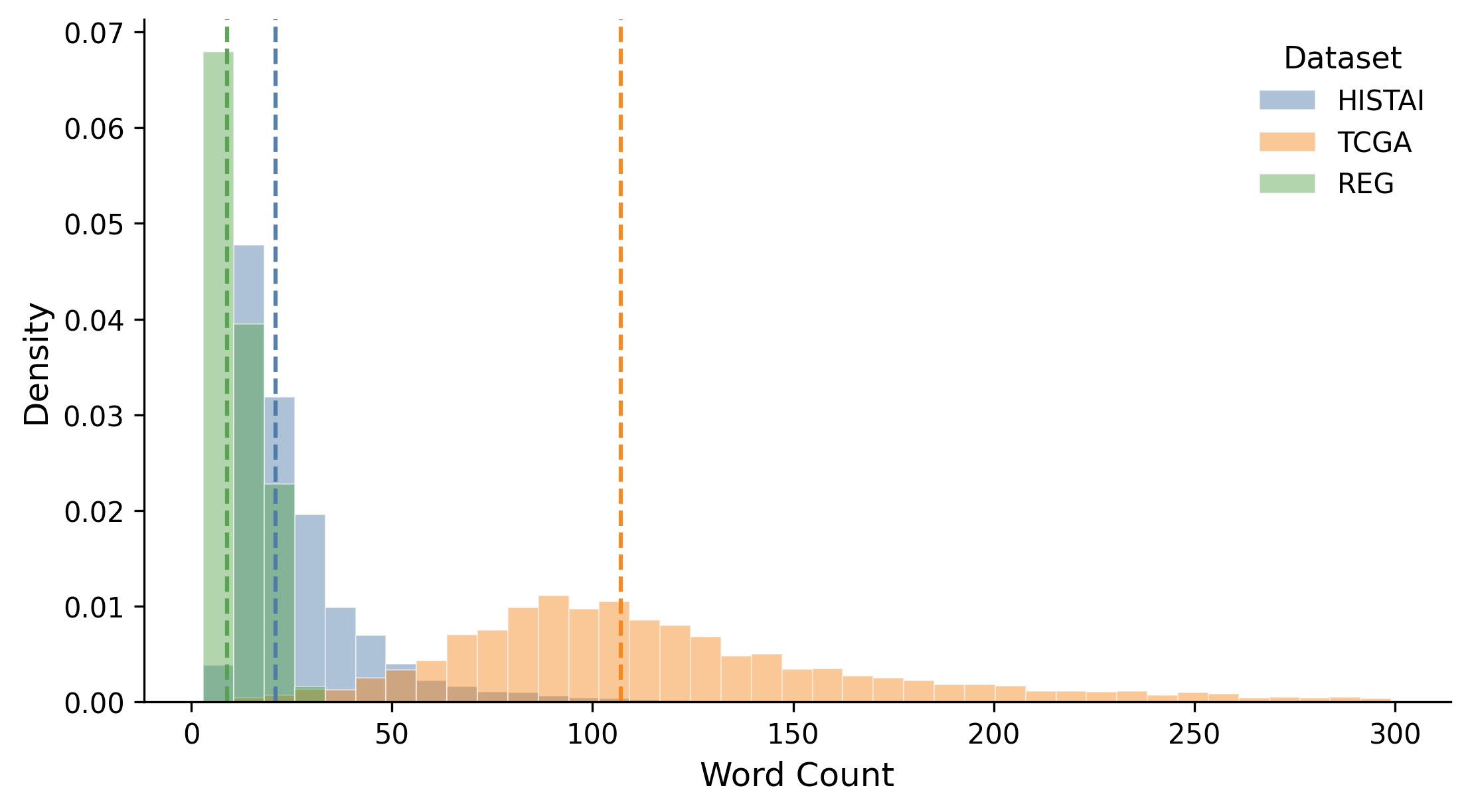}
    \end{minipage}\hfill
    \begin{minipage}[t]{0.5\linewidth}
        \centering
        \includegraphics[width=\linewidth]{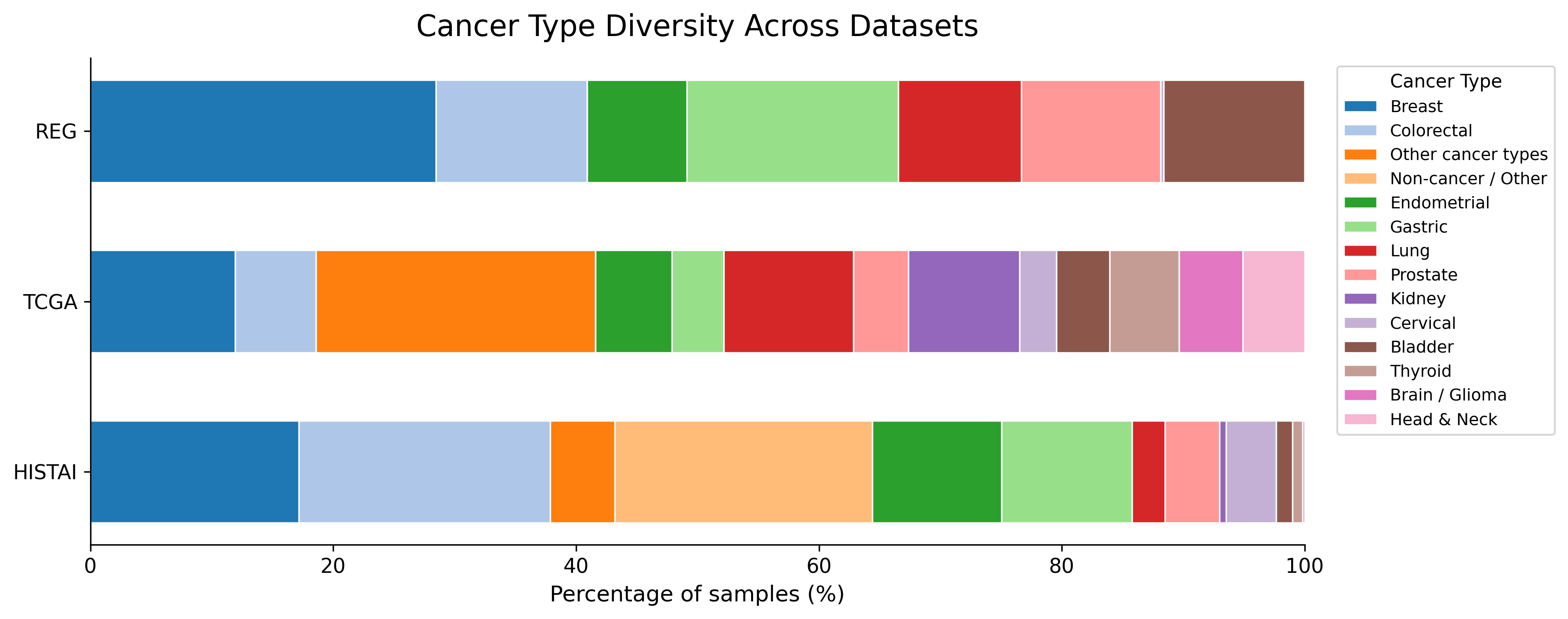}
    \end{minipage}
    \caption{Dataset characteristics across the three pathology report generation benchmarks. Left: distribution of report lengths (word count) across datasets. TCGA exhibits substantially longer and more variable reports, while REG contains shorter and more consistent reports. HistAI shows intermediate characteristics. Dashed lines indicate mean word counts. Right: cancer-type distribution across datasets, highlighting differences in class composition.}
    \label{fig:report_length_distribution}
\end{figure}

\subsection{Datasets}

We evaluate on three pathology report generation datasets with complementary characteristics,
summarized in Table~\ref{tab:dataset_comparison}. The datasets differ in scale, institutional diversity,
report structure, and annotation noise, allowing us to test model behavior across heterogeneous
reporting conditions. As shown in Figure~\ref{fig:report_length_distribution}, the datasets also exhibit distinct report length distributions, reflecting differences in structure and variability.The same figure also shows marked differences in cancer-type composition across datasets, which may influence task difficulty and cross-dataset generalization. Taken together, these distributions highlight that the benchmark captures variation in both report style and disease mix.  \textbf{TCGA (PathText)~\cite{wsi_caption}:}
TCGA consists of WSIs paired with pathology reports derived from The Cancer Genome Atlas and
used in prior WSI-Caption work. Its reports are relatively long and heterogeneous, with higher
annotation noise due to variation in reporting style. \textbf{HistAI~\cite{histai_dataset}:}
HistAI is a large-scale, multi-institutional dataset containing diverse tissue types and moderately structured reports. It captures substantial variability in both image appearance and report language,
making it useful for evaluating robustness under real-world reporting heterogeneity. \textbf{REG 2025~\cite{reg2025_challenge}:} REG 2025 is a benchmark dataset designed for pathology report generation. It contains shorter,
semi-structured reports with higher report consistency and lower annotation noise.

For reproducibility, we release the exact train, validation, and test splits used for all datasets. These
fixed splits ensure that future methods can be evaluated under the same data protocol rather than
under independently constructed partitions.

\subsection{Models and Visual Encoders}

We benchmark four representative pathology report generation methods spanning distinct modeling
strategies: WSI-Caption (MI-Gen)~\cite{wsi_caption}, a multiple-instance generative model; HistGen~\cite{histgen}, a hierarchical
local-global encoder with cross-modal context interaction; BiGen~\cite{bigen}, a retrieval-augmented model that
incorporates external textual knowledge; and SCOUT~\cite{scout}, a context-aware multimodal transformer with
concept-conditioned feature refinement.

To isolate the role of visual representation, each report generation method is evaluated with three
pretrained pathology encoders: CONCHv1.5~\cite{conch}, UNI2-h~\cite{uni}, and H-Optimus-1~\cite{hoptimus1}. All methods are trained and
evaluated using the same data splits, preprocessing pipeline, tokenizer, feature extraction protocol,
and decoding strategy. We reimplement the compared methods in a shared PyTorch Lightning codebase,
which reduces implementation-specific variation across originally separate repositories and enables
controlled comparison of model architecture and encoder choice. All visual features are extracted with TRIDENT~\cite{zhang2025acceleratingdataprocessingbenchmarking, tridentgithub} as explained in section \ref{sec:bench}.

\input{sections/crqs}

\subsection{Benchmarking Framework}

To improve reproducibility and extensibility, we develop a unified benchmarking framework that standardizes data preprocessing, feature extraction, model training, and evaluation across all experiments. As shown in Figure~\ref{fig:benchmark_overview}, the framework is modular and plug-and-play, enabling researchers to integrate new report generation models, evaluate multiple datasets and visual encoders, and compare results under consistent experimental settings. This unified setup reduces implementation variability and, in several cases, produces stronger baseline performance than originally reported, suggesting that consistent preprocessing and training contribute substantially to performance gains. In addition to CRQS, we evaluate models using standard NLG metrics, including BLEU (1--4)~\cite{bleu}, ROUGE-L~\cite{rouge}, and METEOR, and analyze their relationship with CRQS to highlight differences between lexical similarity and clinical correctness.






%% file: sections/crqs.tex
\subsection{Clinical Report Quality Score (CRQS)}
\label{sec:crqs}

\myparagraph{Clinical motivation.}
Existing report-generation benchmarks in pathology commonly rely on NLG metrics such as BLEU, ROUGE, and METEOR. While useful for measuring lexical similarity,
these metrics are poorly suited to evaluating whether a generated report preserves clinically actionable
information. In pathology, small factual errors can be consequential: an omitted margin status, an
incorrect tumor grade, or a hallucinated lymph-node finding can materially alter downstream clinical
interpretation. This motivates an evaluation metric that measures structured clinical fidelity rather
than surface-level textual overlap.

We therefore propose the \emph{Clinical Report Quality Score} (CRQS), a structured and clinically
grounded metric for evaluating pathology report generation. CRQS is motivated by synoptic reporting
practices, such as those recommended by the College of American Pathologists, in which reports are
organized around standardized diagnostic fields. Instead of comparing generated and reference reports
as unstructured text, CRQS organizes extracted clinical fields into clinically meaningful groups, including diagnosis hierarchy, tumor behavior, grade and aggressiveness, invasion patterns, and associated morphologic findings. The hierarchy was designed in consultation with a practicing pathologist and prioritizes findings that are both clinically actionable and inferable from histomorphologic appearance in the whole-slide image (WSI).

Let $R$ denote the reference report and $\hat{R}$ denote the generated report. We define a structured
extraction function
\begin{equation}
    \mathcal{E}(R) = \{(f_i, v_i)\}_{i=1}^{n},
\end{equation}
where $f_i$ denotes a clinical field and $v_i$ denotes its extracted value. The same extraction function
is applied to $\hat{R}$. In practice, $\mathcal{E}$ is implemented using a structured language-model
pipeline that maps free-text reports into a predefined schema of pathology-relevant attributes. Extracted
fields are normalized to canonical terminology where possible, including synonym resolution, negation
handling, and value standardization. A subset of fields, denoted $\mathcal{K}$, is designated as clinically
critical because errors in these fields are more likely to affect diagnostic interpretation or downstream
clinical management. In particular, these key fields emphasize visually grounded diagnostic attributes over metadata heavy fields that may not be directly observable from a single histopathology image. The complete dataset-specific list of extracted clinical fields and key fields is provided in the Appendix (A.7).


For each field $f$, let $v_f$ and $\hat{v}_f$ denote the reference and generated values, respectively,
when available. A field is considered correctly matched if the generated value is clinically equivalent
to the reference value under the normalization and matching rules of the schema. This allows CRQS to
account for clinically equivalent phrasings while penalizing omissions, unsupported findings, and
discordant clinical values.

CRQS is composed of four interpretable components.

\myparagraph{Clinical Fact Coverage.}
Clinical Fact Coverage (CFC) measures the fraction of reference clinical fields that are correctly
recovered in the generated report:
\begin{equation}
    \mathrm{CFC} =
    \frac{
    \sum_{f \in \mathcal{F}_{R}} \mathbf{1}[\hat{v}_f \equiv v_f]
    }{
    |\mathcal{F}_{R}|
    },
\end{equation}
where $\mathcal{F}_{R}$ is the set of clinical fields present in the reference report, and $\equiv$ denotes
clinical equivalence after normalization. CFC captures overall completeness with respect to the
reference report.

\myparagraph{Key Information Recall.}
Key Information Recall (KIR) emphasizes recovery of high-impact diagnostic attributes:
\begin{equation}
    \mathrm{KIR} =
    \frac{
    \sum_{f \in \mathcal{K}_{R}} \mathbf{1}[\hat{v}_f \equiv v_f]
    }{
    |\mathcal{K}_{R}|
    },
\end{equation}
where $\mathcal{K}_{R} = \mathcal{K} \cap \mathcal{F}_{R}$. This term gives additional weight to
clinically critical fields such as diagnosis, grade, stage, margin status, and lymph-node involvement.

\myparagraph{Hallucination Rate.}
Hallucination Rate (HR) measures the proportion of generated clinical facts that are not supported by
the reference report:
\begin{equation}
    \mathrm{HR} =
    \frac{
    \sum_{f \in \mathcal{F}_{\hat{R}}}
    \mathbf{1}[f \notin \mathcal{F}_{R} \ \text{or} \ \hat{v}_f \not\equiv v_f]
    }{
    |\mathcal{F}_{\hat{R}}|
    },
\end{equation}
where $\mathcal{F}_{\hat{R}}$ is the set of clinical fields asserted in the generated report. HR penalizes
unsupported or fabricated findings, which are particularly important in clinical report generation
because fluent but incorrect statements may appear plausible to non-expert evaluators.

\myparagraph{Clinical Discordance Score.}
Clinical Discordance Score (CDS) measures disagreement on fields that are present in both the
reference and generated reports:
\begin{equation}
    \mathrm{CDS} =
    \frac{
    \sum_{f \in \mathcal{F}_{R} \cap \mathcal{F}_{\hat{R}}}
    \mathbf{1}[\hat{v}_f \not\equiv v_f]
    }{
    |\mathcal{F}_{R} \cap \mathcal{F}_{\hat{R}}|
    }.
\end{equation}
Unlike hallucination rate, which penalizes unsupported generated facts, CDS specifically captures
direct contradictions between the generated and reference reports. Examples include assigning the
wrong tumor grade, incorrect stage, or discordant biomarker status.


The final Clinical Report Quality Score combines completeness, critical information recall, hallucination avoidance, and clinical consistency:
\begin{equation}
\mathrm{CRQS}_{\mathrm{raw}}
=
w_1 \mathrm{CFC}
+
w_2 \mathrm{KIR}
+
w_3 (1-\mathrm{HR})
+
w_4 (1-\mathrm{CDS}),
\end{equation}
where $w_1,w_2,w_3,w_4 \geq 0$. Unless otherwise specified, we use
\begin{equation}
w_1 = 0.30, \quad
w_2 = 0.40, \quad
w_3 = 0.20, \quad
w_4 = 0.40.
\end{equation}

These weights place greater emphasis on recovery of clinically critical information while penalizing hallucinated and discordant findings. In particular, CRQS penalizes clinical discordance more heavily than hallucination because contradictory or clinically incorrect findings are typically more dangerous than unsupported but non-conflicting additions. Under this weighting scheme, the theoretical maximum raw score is $0.7$. We therefore normalize the final score as
\begin{equation}
\mathrm{CRQS}
=
\frac{\mathrm{CRQS}_{\mathrm{raw}}}{0.7},
\end{equation}
so that $\mathrm{CRQS} \in [0,1]$, where higher values indicate better clinical report quality.

This decomposition provides two advantages over conventional text-overlap metrics. First, it evaluates
generated reports at the level of clinically meaningful facts rather than lexical similarity. Second, it
yields interpretable sub-scores that distinguish between different failure modes: omission of relevant
information, failure to recover key diagnostic fields, hallucination of unsupported findings, and direct
clinical contradiction. As a result, CRQS enables more clinically faithful comparison of pathology
report generation systems and supports error analysis beyond aggregate language-generation scores.

%% file: sections/experiments.tex
\section{Experiments and Results}

\myparagraph{Experimental Setup.} 
All models are trained using identical preprocessing, feature extraction, and decoding settings, as described in Sec.~\ref{sec:benchmark}. All models are evaluated on held-out test sets. To ensure fairness, identical preprocessing and tokenization are used across methods, visual features are extracted using the same encoder for each experimental setting, and decoding strategies are fixed across models. This standardized protocol ensures that performance differences reflect model capabilities rather than experimental inconsistencies. Further details in Appendix \ref{sec:experimenal_setup}.



\myparagraph{Main Results.} Tables~\ref{tab:tcga_methods}, \ref{tab:histai_methods}, and \ref{tab:reg_methods} summarize method-level performance for each dataset after averaging across the three encoders. Complementarily, Tables~\ref{tab:tcga_encoders}, \ref{tab:histai_encoders}, and \ref{tab:reg_encoders} summarize encoder-level performance for each dataset after averaging across the four methods. This two-view presentation keeps the tables compact while still exposing both model-centric and encoder-centric trends.

\begin{table*}[t]

\centering

\caption{Dataset-wise results averaged across encoders and methods. Best results are in bold}

\label{tab:all_dataset_results}

\begin{subtable}{\textwidth}
\centering
\caption{Method-wise results on TCGA, averaged across encoders.}
\label{tab:tcga_methods}
\resizebox{\textwidth}{!}{
\begin{tabular}{lccccccc}
\toprule
Method & BLEU-1 & BLEU-2 & BLEU-3 & BLEU-4 & METEOR & ROUGE-L & CRQS \\
\midrule
WSI-Caption & 0.3779 & 0.2477 & \textbf{0.1673} & 0.1149 & 0.1529 & 0.2884 & 0.1861 \\
HistGen & 0.3400 & 0.2241 & 0.1519 & 0.1046 & 0.1481 & 0.2840 & 0.1784 \\
BiGen & 0.3640 & 0.2422 & 0.1662 & \textbf{0.1161} & \textbf{0.1535} & \textbf{0.2979} & 0.1882 \\
SCOUT & \textbf{0.3860} & \textbf{0.2500} & 0.1673 & 0.1135 & 0.1508 & 0.2851 & \textbf{0.1930} \\
\bottomrule
\end{tabular}}
\end{subtable}

\vspace{0.6em}

\begin{subtable}{\textwidth}
\centering
\caption{Encoder-wise results on TCGA, averaged across methods.}
\label{tab:tcga_encoders}
\resizebox{\textwidth}{!}{
\begin{tabular}{lccccccc}
\toprule
Encoder & BLEU-1 & BLEU-2 & BLEU-3 & BLEU-4 & METEOR & ROUGE-L & CRQS \\
\midrule
CONCHv1.5 & 0.3507 & 0.2297 & 0.1550 & 0.1062 & 0.1472 & 0.2853 & 0.1750 \\
H-Optimus-1 & 0.3740 & 0.2453 & 0.1660 & 0.1142 & \textbf{0.1534} & 0.2890 & \textbf{0.1923} \\
UNI2-h & \textbf{0.3762} & \textbf{0.2480} & \textbf{0.1685} & \textbf{0.1164} & 0.1533 & \textbf{0.2923} & 0.1919 \\
\bottomrule
\end{tabular}}
\end{subtable}

\vspace{0.6em}

\begin{subtable}{\textwidth}
\centering
\caption{Method-wise results on HistAI, averaged across encoders.}
\label{tab:histai_methods}
\resizebox{\textwidth}{!}{
\begin{tabular}{lccccccc}
\toprule
Method & BLEU-1 & BLEU-2 & BLEU-3 & BLEU-4 & METEOR & ROUGE-L & CRQS \\
\midrule
WSI-Caption & 0.2662 & 0.1884 & 0.1439 & 0.1148 & 0.1465 & 0.3655 & 0.2741 \\
HistGen & 0.2820 & 0.2003 & 0.1541 & 0.1242 & 0.1517 & 0.3640 & 0.2809 \\
BiGen & 0.2682 & 0.1911 & 0.1474 & 0.1189 & 0.1506 & \textbf{0.3696} & \textbf{0.2938} \\
SCOUT & \textbf{0.3143} & \textbf{0.2202} & \textbf{0.1674} & \textbf{0.1336} & \textbf{0.1557} & 0.3629 & 0.2802 \\
\bottomrule
\end{tabular}}
\end{subtable}

\vspace{0.6em}

\begin{subtable}{\textwidth}
\centering
\caption{Encoder-wise results on HistAI, averaged across methods.}
\label{tab:histai_encoders}
\resizebox{\textwidth}{!}{
\begin{tabular}{lccccccc}
\toprule
Encoder & BLEU-1 & BLEU-2 & BLEU-3 & BLEU-4 & METEOR & ROUGE-L & CRQS \\
\midrule
CONCHv1.5 & \textbf{0.2908} & \textbf{0.2038} & 0.1550 & 0.1236 & 0.1513 & 0.3638 & 0.2787 \\
H-Optimus-1 & 0.2844 & 0.2025 & \textbf{0.1557} & \textbf{0.1251} & \textbf{0.1530} & \textbf{0.3694} & \textbf{0.2906} \\
UNI2-h & 0.2728 & 0.1937 & 0.1489 & 0.1198 & 0.1492 & 0.3634 & 0.2775 \\
\bottomrule
\end{tabular}}
\end{subtable}

\vspace{0.6em}

\begin{subtable}{\textwidth}
\centering
\caption{Method-wise results on REG 2025, averaged across encoders.}
\label{tab:reg_methods}
\resizebox{\textwidth}{!}{
\begin{tabular}{lccccccc}
\toprule
Method & BLEU-1 & BLEU-2 & BLEU-3 & BLEU-4 & METEOR & ROUGE-L & CRQS \\
\midrule
WSI-Caption & 0.8316 & 0.7997 & 0.7709 & 0.7458 & 0.5346 & 0.8507 & 0.6892 \\
HistGen & 0.8500 & 0.8177 & 0.7884 & 0.7627 & 0.5479 & 0.8594 & 0.7003 \\
BiGen & 0.8418 & 0.8129 & 0.7864 & 0.7630 & 0.5459 & 0.8593 & \textbf{0.7049} \\
SCOUT & \textbf{0.8580} & \textbf{0.8260} & \textbf{0.7977} & \textbf{0.7731} & \textbf{0.5554} & \textbf{0.8632} & 0.7036 \\
\bottomrule
\end{tabular}}
\end{subtable}

\vspace{0.6em}

\begin{subtable}{\textwidth}
\centering
\caption{Encoder-wise results on REG 2025, averaged across methods.}
\label{tab:reg_encoders}
\resizebox{\textwidth}{!}{
\begin{tabular}{lccccccc}
\toprule
Encoder & BLEU-1 & BLEU-2 & BLEU-3 & BLEU-4 & METEOR & ROUGE-L & CRQS \\
\midrule
CONCHv1.5 & 0.8441 & 0.8130 & 0.7855 & 0.7616 & 0.5447 & 0.8562 & 0.6927 \\
H-Optimus-1 & 0.8445 & 0.8123 & 0.7826 & 0.7565 & 0.5460 & 0.8563 & 0.6986 \\
UNI2-h & \textbf{0.8474} & \textbf{0.8169} & \textbf{0.7894} & \textbf{0.7654} & \textbf{0.5471} & \textbf{0.8620} & \textbf{0.7072} \\
\bottomrule
\end{tabular}}
\end{subtable}

\end{table*}

\begin{table*}[t]
\centering
\caption{Qualitative results highlighting disagreement between METEOR and CRQS.}[4pt]
\label{tab:meteor_crqs_examples}

\normalsize
\renewcommand{\arraystretch}{1.35}
\setlength{\tabcolsep}{4pt}

\begin{tabular}{p{0.37\textwidth} p{0.37\textwidth} c c}
\hline
\textbf{Ground Truth} & \textbf{Prediction} & \textbf{METEOR} & \textbf{CRQS} \\
\hline

Uterine cervix, colposcopic biopsy; High-grade squamous intraepithelial lesion (HSIL; CIN 2)
&
Uterine cervix, colposcopic biopsy; Low-grade squamous intraepithelial lesion (LSIL; CIN 1)
&
0.4264
&
-0.0714 \\

\hline

The lymph node biopsy from the right supraclavicular area corresponds to a reactive lymph node. No evidence of metastatic carcinoma is identified.
&
Morphological and immunophenotypic features correspond to reactive lymph node hyperplasia.
&
0.0114
&
0.9143 \\

\hline
\end{tabular}

\end{table*}

\begin{figure}[t]
    \centering
    \includegraphics[width=\linewidth]{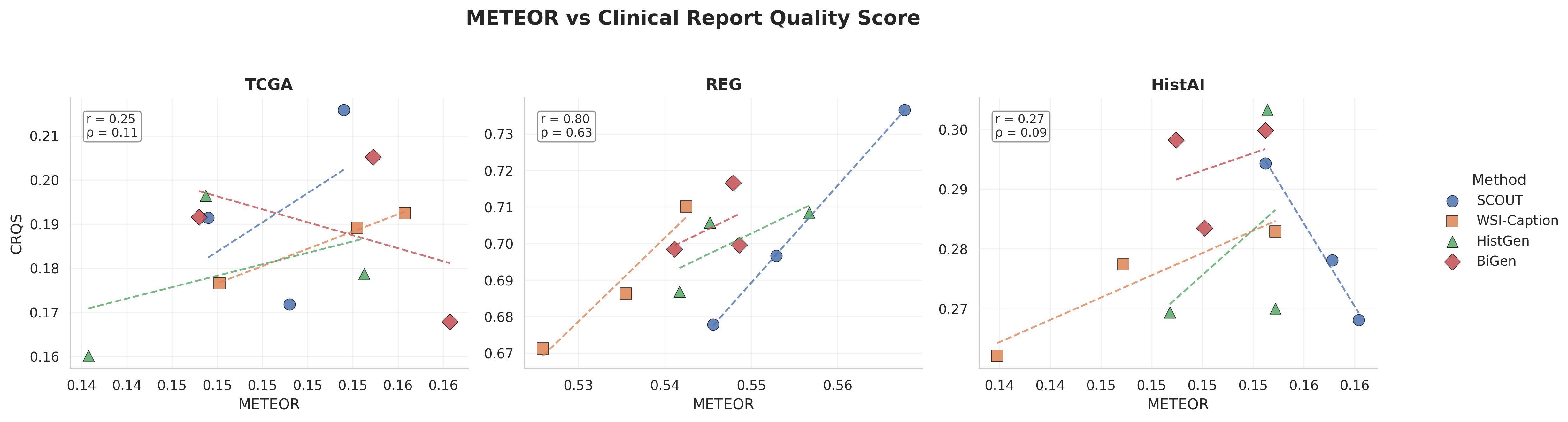}
    \caption{METEOR vs CRQS across datasets.}
    \label{fig:meteor_crqs}
\end{figure}
Across datasets, metric rankings vary by both method and encoder. On REG 2025, SCOUT is the strongest method on the averaged NLG metrics, while BiGen attains the highest averaged CRQS. On TCGA, the rankings are much closer and depend on the metric, with WSI-Caption leading on BLEU-4, SCOUT leading on BLEU-1 and CRQS, and BiGen remaining competitive on ROUGE-L. On HistAI, SCOUT provides the strongest averaged BLEU and METEOR scores, while BiGen attains the highest averaged ROUGE-L and CRQS.

\begin{figure}[t]
    \centering
    \includegraphics[width=\linewidth]{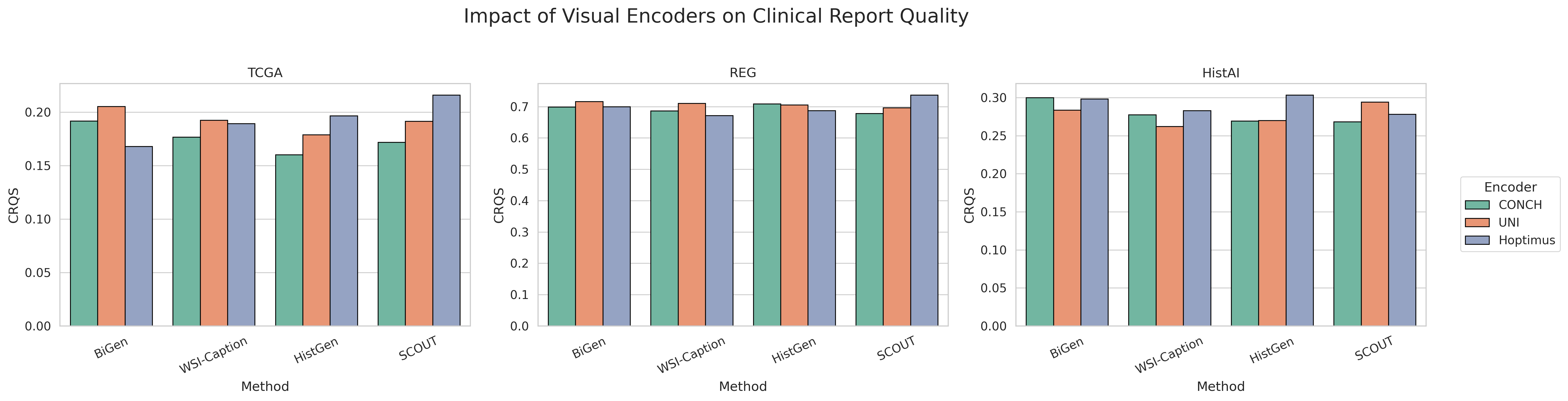}
    \caption{
    Vision encoder sensitivity across TCGA, REG 2025, and HistAI. The grouped bars show Clinical Report Quality Score (CRQS) for four pathology report generation methods using three pathology visual encoders: CONCHv1.5, UNI2-h, and H-Optimus-1. Encoder choice produces method- and dataset-dependent differences in clinically grounded report quality.
    }
    \label{fig:vision_encoder_sensitivity}
\end{figure}

\myparagraph{Encoder Sensitivity.} Figure~\ref{fig:vision_encoder_sensitivity} and the encoder-wise tables show that visual representation choice has a measurable effect on clinically grounded report quality. No single encoder dominates uniformly across every dataset and metric, indicating that encoder sensitivity depends on both the dataset distribution and the downstream generation architecture. On TCGA, H-Optimus-1 yields the strongest averaged BLEU-1 through BLEU-3 and METEOR, while UNI2-h achieves the highest averaged BLEU-4, ROUGE-L, and CRQS. On HistAI, H-Optimus-1 provides the strongest averaged BLEU-3, BLEU-4, METEOR, ROUGE-L, and CRQS, whereas CONCHv1.5 is slightly stronger on BLEU-1 and BLEU-2. On REG 2025, UNI2-h delivers the best averaged performance across all reported NLG metrics and also attains the highest averaged CRQS. These trends reinforce the importance of reporting encoder effects explicitly rather than collapsing them into a single global average.

\myparagraph{Qualitative Analysis.} As shown in Table~\ref{tab:meteor_crqs_examples}, the qualitative examples from the REG and HISTAI datasets respectively, reveal important limitations of traditional NLG metrics for clinical report evaluation. In the REG example, METEOR assigns a moderately high score despite a clinically significant diagnostic error where the prediction incorrectly downgrades the lesion from HSIL/CIN2 to LSIL/CIN1, whereas CRQS correctly assigns a very low score. Conversely, the HISTAI example demonstrates that METEOR can produce an extremely low score even when the generated report preserves the correct benign pathological interpretation using different clinical wording (``reactive lymph node hyperplasia''). In this case, CRQS successfully captures the preserved clinical meaning and assigns a high score. These examples illustrate that lexical overlap metrics primarily evaluate surface text similarity, while CRQS more effectively measures clinically relevant correctness and semantic fidelity.

\myparagraph{Effect of Report Structure on Clinical Fidelity.}
Performance varies substantially across datasets, demonstrating the impact of report structure and clinical heterogeneity on pathology report generation. TCGA, which contains longer and less standardized free-text reports, achieves lower CRQS and substantially higher hallucination rates (HR $\sim$ 0.5), indicating that lexical metrics such as BLEU, ROUGE, and METEOR can overestimate clinical quality by failing to capture clinically meaningful errors. In contrast, REG shows stronger agreement between lexical metrics and CRQS, along with lower hallucination rates, likely due to its shorter and semi-structured reports. HistAI exhibits intermediate behavior. As illustrated in Figure~\ref{fig:meteor_crqs}, the relationship between METEOR and CRQS is inconsistent across datasets, highlighting the need for clinically grounded evaluation metrics such as CRQS and benchmarking across datasets with varying reporting styles and noise levels.

%% file: sections/discussions_and_conclusion.tex
\section{Conclusion}
Our findings carry direct implications for how progress in pathology report generation should be measured. Standard NLG metrics are insufficient proxies for clinical correctness; models with comparable lexical scores can differ substantially in diagnostic accuracy, making such metrics unreliable for guiding development. Clinically grounded evaluation, as instantiated by CRQS, is essential: by decomposing report quality into factual coverage, hallucination rate, and clinical discordance, it surfaces failure modes that surface-level metrics obscure. Equally important is the practice of benchmarking across diverse datasets, encoders, and architectures — single-setting evaluations risk overstating generalization and masking dataset-specific confounds. Taken together, these findings argue for a shift in evaluation norms: clinical correctness, not lexical similarity, should serve as the primary criterion for assessing pathology report generation systems. Overall, the PathReportEval benchmark, public framework, and evaluation metric provide a foundation for more reliable comparison and future development of clinically faithful pathology report generation models.

%% file: sections/appendix.tex
\appendix
\section{Technical appendices and supplementary material}

\subsection{Benchmark and Evaluation Framework}
\label{sec:bench}
TRIDENT provides a standardized pipeline that performs tissue segmentation, patch coordinate extraction at a target magnification, and encoder-specific feature computation from the resulting tissue regions. In our workflow, WSIs are first segmented to remove background regions, after which tissue patches are generated at 20$\times$ magnification before feature encoding. We use TRIDENT to produce the encoder inputs for CONCHv1.5~\cite{conch}, TITAN~\cite{titan}, H-Optimus-1~\cite{hoptimus1}, and UNI2-h~\cite{uni} under a common processing stack, which keeps segmentation, patch extraction, and feature serialization consistent across experiments. Following the TRIDENT model registry, CONCHv1.5 uses 512\,$\times$\,512 patches at 20$\times$, whereas other encoders use their corresponding encoder-specific patch configurations within the same framework~\cite{tridentgithub}. This standardized extraction stage reduces variation introduced by heterogeneous preprocessing pipelines and makes the downstream comparison of report generation models more controlled.

\subsection{Experimental Setup}
\label{sec:experimenal_setup}

Visual features are extracted with the TRIDENT whole-slide processing framework using a shared segmentation and 20$\times$ patch-extraction pipeline before encoder-specific embedding computation. To ensure controlled comparison, all methods are reimplemented within a common PyTorch Lightning framework using the same tokenizer and standardized data-cleaning pipeline. Experiments are conducted on NVIDIA H100 and H200 GPUs with either 4 or 8 nodes. The available memory per gpu is 32 GB and and a batch size of 1 is used for all methods. For reproducibility, we make the dataset splits, benchmarking pipeline, and evaluation code publicly available through our GitHub repository. This unified setup reduces implementation-specific variance and likely contributes to the stronger performance we observe for WSI-Caption, HistGen, and BiGen relative to their originally reported results.

Performance is evaluated using both standard natural language generation (NLG) metrics (BLEU-1, BLEU-2, BLEU-3, BLEU-4, ROUGE-L, and METEOR) and the proposed clinically grounded metric (CRQS).

\subsection{Experiments}

This appendix reports the full per-method and per-encoder results from the final experimental runs for each dataset. Each table lists the individual model--encoder combinations before averaging and includes both standard NLG metrics and CRQS. Within each dataset table, the best score in each metric column is shown in bold.

\subsubsection{TCGA Detailed Results}

\begin{table}[h]
\centering
\caption{Detailed results on TCGA for each method--encoder combination.}
\resizebox{\textwidth}{!}{%
\begin{tabular}{llccccccc}
\toprule
Method & Encoder & BLEU-1 & BLEU-2 & BLEU-3 & BLEU-4 & METEOR & ROUGE-L & CRQS \\
\midrule
WSI-Caption & CONCHv1.5 & 0.3558 & 0.2333 & 0.1569 & 0.1071 & 0.1481 & 0.2863 & 0.1766 \\
WSI-Caption & H-Optimus-1 & 0.3832 & 0.2521 & 0.1707 & 0.1177 & 0.1542 & 0.2907 & 0.1892 \\
WSI-Caption & UNI2-h & 0.3948 & \textbf{0.2577} & \textbf{0.1744} & 0.1198 & 0.1563 & 0.2881 & 0.1924 \\
HistGen & CONCHv1.5 & 0.3313 & 0.2148 & 0.1433 & 0.0971 & 0.1423 & 0.2749 & 0.1601 \\
HistGen & H-Optimus-1 & 0.3313 & 0.2201 & 0.1503 & 0.1040 & 0.1475 & 0.2840 & 0.1964 \\
HistGen & UNI2-h & 0.3573 & 0.2375 & 0.1621 & 0.1128 & 0.1545 & 0.2931 & 0.1787 \\
BiGen & CONCHv1.5 & 0.3278 & 0.2197 & 0.1518 & 0.1070 & 0.1472 & 0.2966 & 0.1916 \\
BiGen & H-Optimus-1 & 0.3802 & 0.2521 & 0.1725 & 0.1198 & \textbf{0.1583} & 0.2984 & 0.1679 \\
BiGen & UNI2-h & 0.3841 & 0.2548 & 0.1742 & \textbf{0.1214} & 0.1549 & \textbf{0.2986} & 0.2052 \\
SCOUT & CONCHv1.5 & 0.3878 & 0.2509 & 0.1679 & 0.1138 & 0.1512 & 0.2833 & 0.1718 \\
SCOUT & H-Optimus-1 & \textbf{0.4014} & 0.2571 & 0.1707 & 0.1151 & 0.1536 & 0.2829 & \textbf{0.2158} \\
SCOUT & UNI2-h & 0.3687 & 0.2421 & 0.1632 & 0.1115 & 0.1476 & 0.2892 & 0.1914 \\
\bottomrule
\end{tabular}%
}
\end{table}

\subsubsection{HistAI Detailed Results}

\begin{table}[h]
\centering
\caption{Detailed results on HistAI for each method--encoder combination.}
\resizebox{\textwidth}{!}{%
\begin{tabular}{llccccccc}
\toprule
Method & Encoder & BLEU-1 & BLEU-2 & BLEU-3 & BLEU-4 & METEOR & ROUGE-L & CRQS \\
\midrule
WSI-Caption & CONCHv1.5 & 0.2598 & 0.1841 & 0.1401 & 0.1117 & 0.1461 & 0.3707 & 0.2774 \\
WSI-Caption & H-Optimus-1 & 0.2897 & 0.2067 & 0.1591 & 0.1276 & 0.1536 & 0.3688 & 0.2829 \\
WSI-Caption & UNI2-h & 0.2492 & 0.1743 & 0.1325 & 0.1051 & 0.1399 & 0.3571 & 0.2621 \\
HistGen & CONCHv1.5 & 0.2729 & 0.1928 & 0.1477 & 0.1184 & 0.1484 & 0.3599 & 0.2694 \\
HistGen & H-Optimus-1 & 0.2824 & 0.2012 & 0.1553 & 0.1257 & 0.1532 & 0.3693 & \textbf{0.3032} \\
HistGen & UNI2-h & 0.2906 & 0.2069 & 0.1594 & 0.1286 & 0.1536 & 0.3628 & 0.2700 \\
BiGen & CONCHv1.5 & 0.2707 & 0.1931 & 0.1493 & 0.1206 & 0.1531 & \textbf{0.3721} & \textbf{0.2998} \\
BiGen & H-Optimus-1 & 0.2673 & 0.1905 & 0.1468 & 0.1182 & 0.1487 & 0.3703 & 0.2982 \\
BiGen & UNI2-h & 0.2667 & 0.1898 & 0.1461 & 0.1178 & 0.1501 & 0.3665 & 0.2835 \\
SCOUT & CONCHv1.5 & \textbf{0.3598} & \textbf{0.2453} & \textbf{0.1830} & \textbf{0.1437} & \textbf{0.1577} & 0.3524 & 0.2681 \\
SCOUT & H-Optimus-1 & 0.2984 & 0.2115 & 0.1614 & 0.1291 & 0.1564 & 0.3691 & 0.2781 \\
SCOUT & UNI2-h & 0.2846 & 0.2038 & 0.1578 & 0.1279 & 0.1531 & 0.3672 & 0.2943 \\
\bottomrule
\end{tabular}%
}
\end{table}

\subsubsection{REG 2025 Detailed Results}

\begin{table}[h]
\centering
\caption{Detailed results on REG 2025 for each method--encoder combination.}
\resizebox{\textwidth}{!}{%
\begin{tabular}{llccccccc}
\toprule
Method & Encoder & BLEU-1 & BLEU-2 & BLEU-3 & BLEU-4 & METEOR & ROUGE-L & CRQS \\
\midrule
WSI-Caption & CONCHv1.5 & 0.8372 & 0.8035 & 0.7741 & 0.7488 & 0.5355 & 0.8488 & 0.6863 \\
WSI-Caption & H-Optimus-1 & 0.8221 & 0.7881 & 0.7561 & 0.7281 & 0.5259 & 0.8428 & 0.6713 \\
WSI-Caption & UNI2-h & 0.8356 & 0.8075 & 0.7824 & 0.7606 & 0.5425 & 0.8605 & 0.7101 \\
HistGen & CONCHv1.5 & 0.8601 & 0.8296 & 0.8032 & \textbf{0.7802} & 0.5567 & 0.8643 & 0.7084 \\
HistGen & H-Optimus-1 & 0.8395 & 0.8067 & 0.7762 & 0.7494 & 0.5417 & 0.8525 & 0.6868 \\
HistGen & UNI2-h & 0.8505 & 0.8169 & 0.7858 & 0.7585 & 0.5452 & 0.8615 & 0.7057 \\
BiGen & CONCHv1.5 & 0.8322 & 0.8032 & 0.7763 & 0.7528 & 0.5411 & 0.8571 & 0.6985 \\
BiGen & H-Optimus-1 & 0.8512 & 0.8208 & 0.7931 & 0.7689 & 0.5486 & 0.8561 & 0.6996 \\
BiGen & UNI2-h & 0.8419 & 0.8147 & 0.7897 & 0.7673 & 0.5479 & 0.8648 & 0.7166 \\
SCOUT & CONCHv1.5 & 0.8471 & 0.8157 & 0.7883 & 0.7645 & 0.5456 & 0.8545 & 0.6778 \\
SCOUT & H-Optimus-1 & \textbf{0.8651} & \textbf{0.8337} & \textbf{0.8050} & 0.7797 & \textbf{0.5677} & \textbf{0.8739} & \textbf{0.7365} \\
SCOUT & UNI2-h & 0.8617 & 0.8286 & 0.7997 & 0.7752 & 0.5529 & 0.8611 & 0.6966 \\
\bottomrule
\end{tabular}%
}
\end{table}

\subsection{Vision Encoder Sensitivity}

\begin{figure}[H]
    \centering
    \includegraphics[width=\linewidth]{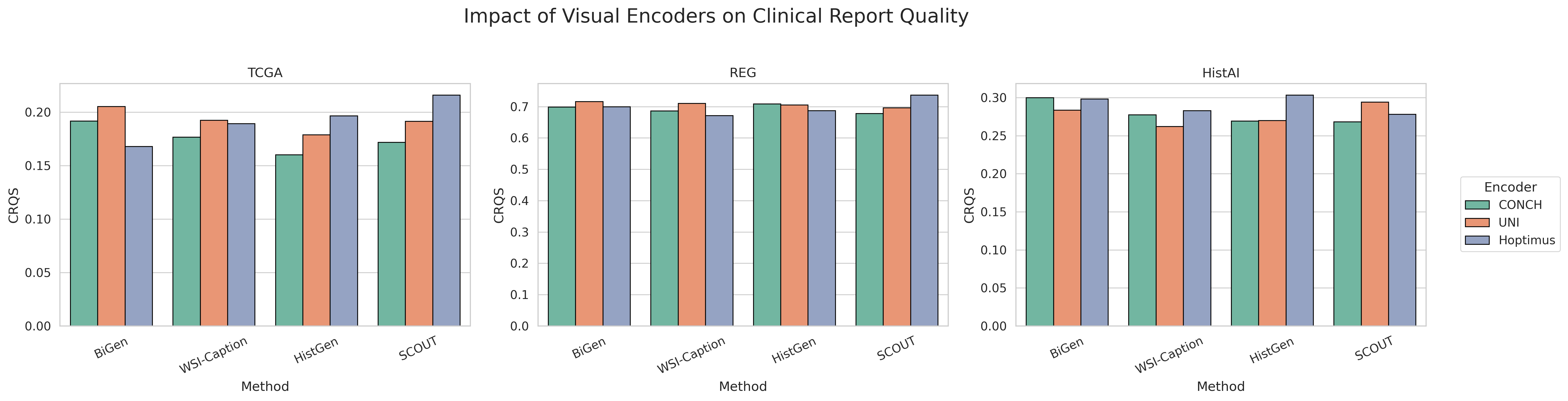}
    \caption{
    Impact of different vision encoders on Clinical Report Quality Score (CRQS) across the TCGA, REG, and HistAI datasets. 
    The comparison evaluates four pathology report generation methods (BiGen, WSI-Caption, HistGen, and SCOUT) using three visual encoders: CONCH, UNI, and Hoptimus. 
    Results show that encoder selection comparably affects clinically grounded report quality, with dataset-specific performance variations observed across methods.
    }
    \label{fig:vision_encoder_sensitivity_app}
\end{figure}

\subsection{Limitations}
\label{limitations}
Despite its contributions, this study has several limitations. First, the proposed CRQS metric relies on structured extraction of clinical fields using a language model-based pipeline. Although these extracted fields are verified by expert pathologists, the extraction process may still introduce errors or inconsistencies, particularly for ambiguous or complex cases.

Second, CRQS focuses on predefined clinical attributes and may not capture all aspects of report quality, such as nuanced descriptive language or rare pathological findings. Extending the metric to incorporate a broader range of clinical concepts remains an important direction for future work.

Third, our benchmark is limited to three datasets and a fixed set of models and encoders. While these choices cover a range of scenarios, additional datasets and emerging architectures may further enrich the evaluation landscape.

\subsection{Broader Impacts}
\label{broader}

Automated pathology report generation has the potential to improve clinical workflow efficiency and reduce documentation burden. However, inaccurate or misleading reports could have serious consequences in medical settings. By emphasizing clinically grounded evaluation, this work aims to promote the development of more reliable and trustworthy systems.

At the same time, the use of automated extraction methods and machine learning models raises concerns about bias, generalization, and interpretability. Ensuring that such systems are rigorously validated and used as decision-support tools, rather than replacements for expert judgment, is critical for safe deployment.

\subsection{Clinical Fields Extraction for each dataset}
\label{sec:crqs_field_extraction}
The following prompt was used to guide the large language model (LLM) in generating dataset-specific clinical fields and key fields for the Clinical Report Quality Score (CRQS) framework. The prompt was designed to enforce clinically meaningful, pathology-oriented extraction schemas rather than using a universal reporting template.

\begin{quote}
\small
\begin{verbatim}
I am building a dataset-specific CRQS pipeline for pathology
report evaluation.

CRQS compares extracted clinical facts from:
1. ground-truth pathology report
2. predicted/generated pathology report

Do NOT use a universal schema. Build a dataset-specific schema
based on the dataset structure and the pathology content.

Important clinical constraint:
The fields should prioritize facts that are clinically important
AND inferable from the histology slide/WSI whenever possible.
Do not over-prioritize metadata fields such as organ site,
specimen type, laterality, procedure type, margin status,
lymph node status, or tumor stage unless they are clearly
present and visually inferable in this dataset.

Use the pathologist-informed hierarchy:
1. Diagnosis hierarchy
   - benign_vs_malignant
   - diagnosis
   - primary_vs_metastatic
   - lineage
   - histologic_type

2. Tumor behavior
   - in_situ_vs_invasive
   - in_situ_component

3. Grade / aggressiveness
   - tumor_grade
   - differentiation
   - dataset-specific grades such as Gleason score,
     WHO grade, nuclear grade, mitotic score,
     tubule score

4. Invasion hierarchy
   - invasion_status
   - invasion_depth
   - organ-specific invasion
   - lymphovascular_invasion
   - perineural_invasion

5. Premalignant / epithelial change
   - dysplasia_grade
   - epithelial_atypia

6. Associated morphology
   - necrosis
   - inflammation_status
   - calcification
   - tumor_volume or tumor_extent if visually meaningful

For this dataset, do the following step by step:

Step 1: Read and summarize the dataset.
Explain:
- file structure
- keys/columns
- whether it has target only or target + prediction
- number of cases if visible
- report style
- organ/disease scope
- examples of report text

Step 2: Identify all clinically relevant fields for this dataset.
Create a single hierarchical flat list:
CLINICAL_FIELDS = [...]
The list should be ordered so the hierarchy is obvious when
reading top to bottom.

Step 3: Select top 10 KEY_FIELDS.
These must be:
- clinically high-impact
- commonly present in this dataset
- inferable from WSI/histology image when possible
- not mostly metadata

Step 4: Explain why each key field was selected.

Step 5: Write/update config.py.
Include:
- CLINICAL_FIELDS
- KEY_FIELDS
- CRQS_WEIGHTS
- STRICT_FIELDS
- NUMERIC_FIELDS

Use raw CRQS weights:
CRQS = 0.3*CFC + 0.4*KIR - 0.2*HR - 0.4*CDS

Also include normalized CRQS:
CRQS_norm = CRQS_raw / 0.7

Step 6: Write extract_fields.py.
Make diagnosis/histology extraction data-driven when possible:
- learn vocabulary from dataset reports
- use stable regex for structured fields such as grade,
  score, invasion, necrosis, tumor size/volume
- avoid manually hard-coding every diagnosis unless needed
  for normalization
- include smoke-test examples

Step 7: Write test_extract_fields.py.
The smoke test should:
- load vocabulary
- run several representative reports
- check expected extracted fields
- print PASS/FAIL

Step 8: Write compute_metrics.py.
Implement:
- CFC = correct target clinical fields /
        target clinical fields present
- KIR = correct key fields /
        target key fields present
- HR = predicted unsupported fields /
       predicted fields
- CDS = discordant comparable fields /
        comparable fields
- CRQS_raw
- CRQS_norm

Step 9: Write run_crqs.py.
It should:
- load dataset
- learn/save/load vocabulary
- extract target and prediction facts
- compute metrics per case
- save per-case CSV/JSON
- save summary CSV/JSON
- print summary

Step 10: Add a short explanation of how to run:
python src/extract_fields.py
python src/test_extract_fields.py
python src/compute_metrics.py
python src/run_crqs.py --input data/<dataset>.json
                         --output_dir outputs

Important:
Use dataset-specific fields only.
Do not assume fields that are not present in the dataset.
\end{verbatim}
\end{quote}

\begin{table}[H]
\centering
\caption{Clinical Fields for REG Dataset}
\label{tab:reg_clinical_fields}
\renewcommand{\arraystretch}{1.2}
\begin{tabular}{|p{4cm}|p{10cm}|}
\hline
\textbf{Field} & \textbf{Description} \\
\hline
organ\_site & Tissue/organ sampled \\
\hline
specimen\_type & Procedure/sample type \\
\hline
diagnosis & Main pathologic diagnosis \\
\hline
histologic\_type & Tumor subtype \\
\hline
tumor\_grade & Degree of differentiation/aggressiveness \\
\hline
gleason\_score & Prostate grading score \\
\hline
grade\_group & ISUP prostate group \\
\hline
tumor\_volume & Estimated involvement percentage \\
\hline
differentiation & Well/moderate/poor differentiation \\
\hline
invasion\_depth & Extent of invasion \\
\hline
in\_situ\_component & Presence of DCIS/CIS/LSIL etc. \\
\hline
nuclear\_grade & Cytologic grade \\
\hline
mitotic\_score & Mitotic count component \\
\hline
tubule\_score & Breast Nottingham component \\
\hline
necrosis & Tumor necrosis present/absent \\
\hline
calcification & Microcalcification presence \\
\hline
dysplasia\_grade & Low/high grade dysplasia \\
\hline
muscle\_proper\_status & Status of muscularis propria involvement/presence \\
\hline
\end{tabular}
\end{table}

\begin{table}[H]
\centering
\caption{Key Fields for REG Dataset}
\label{tab:reg_key_fields}
\renewcommand{\arraystretch}{1.2}
\begin{tabular}{|p{6cm}|p{8cm}|}
\hline
\textbf{Key Field} & \textbf{Clinical Importance} \\
\hline
diagnosis & Primary pathologic diagnosis driving treatment decisions \\
\hline
histologic\_type & Defines tumor subtype and disease classification \\
\hline
tumor\_grade & Indicates aggressiveness and prognosis \\
\hline
gleason\_score & Critical grading system for prostate carcinoma \\
\hline
invasion\_depth & Determines extent of local tumor invasion \\
\hline
\end{tabular}
\end{table}

\begin{table}[H]
\centering
\caption{Clinical Fields for HistAI Dataset}
\label{tab:histai_clinical_fields}
\renewcommand{\arraystretch}{1.2}
\begin{tabular}{|p{5cm}|p{9cm}|}
\hline
\textbf{Field} & \textbf{Description} \\
\hline

\multicolumn{2}{|c|}{\textbf{1. Diagnosis Hierarchy}} \\
\hline
benign\_vs\_malignant & Benign versus malignant disease classification \\
\hline
diagnosis & Primary pathologic diagnosis \\
\hline
primary\_vs\_metastatic & Distinguishes primary tumor from metastatic disease \\
\hline
lineage & Cellular/tissue lineage of tumor \\
\hline
histologic\_type & Histologic subtype of lesion/tumor \\
\hline

\multicolumn{2}{|c|}{\textbf{2. Tumor Behavior}} \\
\hline
in\_situ\_vs\_invasive & Whether lesion is in-situ or invasive \\
\hline
in\_situ\_component & Presence of in-situ neoplastic component \\
\hline

\multicolumn{2}{|c|}{\textbf{3. Grade / Aggressiveness}} \\
\hline
tumor\_grade & Overall tumor grade/aggressiveness \\
\hline
differentiation & Degree of differentiation \\
\hline
gleason\_score & Prostate Gleason grading score \\
\hline
grade\_group & ISUP prostate grade group \\
\hline
nuclear\_grade & Cytologic/nuclear atypia grade \\
\hline
mitotic\_score & Mitotic activity score/count \\
\hline

\multicolumn{2}{|c|}{\textbf{4. Invasion Hierarchy}} \\
\hline
invasion\_status & Presence or absence of invasion \\
\hline
invasion\_depth & Extent/depth of tissue invasion \\
\hline
lymphovascular\_invasion & Tumor invasion into lymphatic or vascular spaces \\
\hline
perineural\_invasion & Tumor invasion involving nerves \\
\hline
extraprostatic\_extension & Tumor extension beyond prostate capsule \\
\hline

\multicolumn{2}{|c|}{\textbf{5. Premalignant / Epithelial Change}} \\
\hline
dysplasia\_grade & Severity of epithelial dysplasia \\
\hline
epithelial\_atypia & Presence of epithelial atypia \\
\hline

\multicolumn{2}{|c|}{\textbf{6. Associated Morphology}} \\
\hline
necrosis & Presence of tumor necrosis \\
\hline
inflammation\_status & Presence of inflammatory changes \\
\hline
inflammation\_activity & Degree/activity of inflammation \\
\hline
calcification & Presence of calcifications \\
\hline
tumor\_extent & Extent of tumor involvement \\
\hline
tumor\_volume & Tumor burden or percentage involvement \\
\hline

\multicolumn{2}{|c|}{\textbf{7. GI-Specific Mucosal Pathology}} \\
\hline
atrophy & Mucosal or glandular atrophy \\
\hline
intestinal\_metaplasia & Presence of intestinal metaplasia \\
\hline
metaplasia\_type & Type/subtype of metaplasia \\
\hline
hpylori\_status & Helicobacter pylori infection status \\
\hline
gastritis\_type & Histologic subtype of gastritis \\
\hline
olga\_stage & OLGA gastritis staging \\
\hline
olgim\_stage & OLGIM intestinal metaplasia staging \\
\hline

\multicolumn{2}{|c|}{\textbf{8. Polyp / Adenoma Pathology}} \\
\hline
polyp\_type & Histologic type of polyp \\
\hline
serrated\_lesion & Presence of serrated lesion morphology \\
\hline
adenoma\_type & Histologic subtype of adenoma \\
\hline

\multicolumn{2}{|c|}{\textbf{9. Multifocal / Bilateral Disease}} \\
\hline
multifocality & Presence of multiple tumor foci \\
\hline
bilateral\_involvement & Bilateral organ involvement \\
\hline

\end{tabular}
\end{table}

\begin{table}[H]
\centering
\caption{Key Fields for HistAI Dataset}
\label{tab:histai_key_fields}
\renewcommand{\arraystretch}{1.2}
\begin{tabular}{|p{5cm}|p{9cm}|}
\hline
\textbf{Key Field} & \textbf{Clinical Importance} \\
\hline
benign\_vs\_malignant & Fundamental distinction guiding diagnosis and treatment \\
\hline
diagnosis & Primary pathologic diagnosis with highest clinical relevance \\
\hline
histologic\_type & Defines tumor subtype and biologic behavior \\
\hline
in\_situ\_vs\_invasive & Determines invasive potential and prognosis \\
\hline
tumor\_grade & Reflects tumor aggressiveness and expected clinical course \\
\hline
gleason\_score & Critical prognostic grading system for prostate carcinoma \\
\hline
dysplasia\_grade & Indicates severity of premalignant epithelial abnormality \\
\hline
invasion\_status & Captures whether tumor has invaded surrounding tissue \\
\hline
intestinal\_metaplasia & Important premalignant gastric mucosal alteration \\
\hline
inflammation\_status & Common histologic finding relevant to disease activity \\
\hline
\end{tabular}
\end{table}

\begin{table}[H]
\centering
\caption{Clinical Fields for TCGA Dataset}
\label{tab:tcga_clinical_fields}
\renewcommand{\arraystretch}{1.2}
\begin{tabular}{|p{5cm}|p{9cm}|}
\hline
\textbf{Field} & \textbf{Description} \\
\hline

\multicolumn{2}{|c|}{\textbf{1. Diagnosis Hierarchy}} \\
\hline
benign\_vs\_malignant & Benign versus malignant disease classification \\
\hline
diagnosis & Primary pathologic diagnosis \\
\hline
primary\_vs\_metastatic & Distinguishes primary tumor from metastatic disease \\
\hline
lineage & Cellular or tissue lineage of tumor \\
\hline
histologic\_type & Histologic subtype of tumor or lesion \\
\hline

\multicolumn{2}{|c|}{\textbf{2. Tumor Behavior}} \\
\hline
in\_situ\_vs\_invasive & Whether lesion is in-situ or invasive \\
\hline
in\_situ\_component & Presence of an in-situ neoplastic component \\
\hline
tumor\_focality & Unifocal versus multifocal tumor distribution \\
\hline

\multicolumn{2}{|c|}{\textbf{3. Grade / Aggressiveness}} \\
\hline
tumor\_grade & Overall tumor grade or aggressiveness \\
\hline
differentiation & Degree of tumor differentiation \\
\hline

\multicolumn{2}{|c|}{\textbf{4. Invasion Hierarchy}} \\
\hline
invasion\_status & Presence or absence of invasion \\
\hline
lymphovascular\_invasion & Tumor invasion into lymphatic or vascular spaces \\
\hline
perineural\_invasion & Tumor invasion involving nerves \\
\hline
capsular\_invasion & Tumor invasion through or into capsule \\
\hline
organ\_specific\_invasion & Dataset-specific invasion into adjacent organ structures \\
\hline

\multicolumn{2}{|c|}{\textbf{5. Spread / Involvement}} \\
\hline
metastatic\_involvement & Tumor involvement of distant or secondary sites \\
\hline
lymph\_node\_involvement & Presence of tumor involvement in lymph nodes \\
\hline
multifocal\_involvement & Multiple sites or foci involved by disease \\
\hline

\multicolumn{2}{|c|}{\textbf{6. Premalignant / Epithelial Change}} \\
\hline
dysplasia\_grade & Severity of epithelial dysplasia \\
\hline
epithelial\_atypia & Presence of epithelial atypia \\
\hline
carcinoma\_in\_situ & Presence of carcinoma in situ \\
\hline
intratubular\_germ\_cell\_neoplasia & Presence of intratubular germ cell neoplasia \\
\hline

\multicolumn{2}{|c|}{\textbf{7. Associated Morphology}} \\
\hline
necrosis & Presence of tumor necrosis \\
\hline
inflammation\_status & Presence of inflammatory changes \\
\hline
calcification & Presence of calcifications \\
\hline
psammoma\_bodies & Presence of psammoma bodies \\
\hline
hemorrhage & Presence of hemorrhage \\
\hline
papillary\_features & Presence of papillary architectural features \\
\hline
cystic\_change & Presence of cystic change \\
\hline

\multicolumn{2}{|c|}{\textbf{8. Tumor Extent / Burden}} \\
\hline
tumor\_size & Tumor dimension or size when reported \\
\hline

\multicolumn{2}{|c|}{\textbf{9. Contextual / Lower-Priority Metadata}} \\
\hline
organ\_site & Anatomical organ or tissue site \\
\hline
specimen\_type & Type of specimen or sampling procedure \\
\hline
margin\_status & Presence or absence of tumor at surgical margins \\
\hline

\end{tabular}
\end{table}

\begin{table}[H]
\centering
\caption{Key Fields for TCGA Dataset}
\label{tab:tcga_key_fields}
\renewcommand{\arraystretch}{1.2}
\begin{tabular}{|p{5cm}|p{9cm}|}
\hline
\textbf{Key Field} & \textbf{Clinical Importance} \\
\hline
benign\_vs\_malignant & Fundamental distinction between benign and malignant disease \\
\hline
diagnosis & Primary pathologic diagnosis with direct clinical relevance \\
\hline
histologic\_type & Defines tumor subtype and disease classification \\
\hline
tumor\_grade & Captures tumor aggressiveness and prognostic severity \\
\hline
differentiation & Indicates biologic behavior and degree of maturation \\
\hline
invasion\_status & Determines whether tumor shows invasive behavior \\
\hline
lymphovascular\_invasion & High-impact feature associated with metastatic risk \\
\hline
metastatic\_involvement & Indicates spread beyond the primary site \\
\hline
necrosis & Common histologic marker of aggressive tumor biology \\
\hline
\end{tabular}
\end{table}